# A Branch-and-Bound Algorithm for MDL Learning Bayesian Networks

Jin Tian
Cognitive Systems Laboratory
Computer Science Department
University of California, Los Angeles, CA 90024
jtian@cs.ucla.edu

## Abstract

This paper extends the work in [Suzuki, 1996] and presents an efficient depth-first branch-and-bound algorithm for learning Bayesian network structures, based on the minimum description length (MDL) principle, for a given (consistent) variable ordering. The algorithm exhaustively searches through all network structures and guarantees to find the network with the best MDL score. Preliminary experiments show that the algorithm is efficient, and that the time complexity grows slowly with the sample size. The algorithm is useful for empirically studying both the performance of suboptimal heuristic search algorithms and the adequacy of the MDL principle in learning Bayesian networks.

## 1 Introduction

Learning the structures of Bayesian networks from data has become an active research area in recent years [Heckerman, 1995, Buntine, 1996]. One approach to this problem is to turn it into an optimization exercise. A scoring function is introduced that evaluates a network with respect to the training data and outputs a number that reflects how well the network scores relative to the available data. We then search through possible network structures for the best scored network and take it as the network learned from the data.

Different scoring functions have been applied such as Bayesian [Cooper and Herskovits, 1992, Heckerman et al., 1995] and minimum description length (MDL) [Bouckaert, 1994a, Lam and Bacchus, 1994, Suzuki, 1996] scoring functions, and various search algorithms have been developed. Since, in general, the search problem is NP-hard [Chickering, 1996], most algorithms use heuristic search methods. Additionally, to reduce the search complexity, some algorithms require as input a strict ordering of variables.

Assuming a consistent variable ordering is given, [Suzuki, 1996] developed a branch-and-bound algorithm using MDL scoring function, which exhaustively searches through all network structures and guarantees to find the best scored network. In this paper, we extend Suzuki's work and present a more efficient branch-and-bound algorithm making use of the special properties of the MDL scoring function. We find bounds for the MDL scores of complex network structures using inequalities from information theory. A greedy search is applied before the branch-and-bound procedure to speed up the pruning process. We show that the MDL scoring function will not select networks with parent sets containing more than $\lfloor \log \frac{2N}{\log N} \rfloor$ variables where $N$ is the sample size. Preliminary test results show that the proposed algorithm is efficient, and that the time complexity grows slowly with respect to the sample size. The algorithm is useful for studying the performance of suboptimal heuristic search algorithms and it paves the way for empirically investigating the MDL principle in learning Bayesian networks.

The paper is organized as follows. Section 2 briefly describes the procedure of learning Bayesian networks from data using the MDL principle. Section 3 reviews some related previous work. Section 4 formally defines the search space of our problem. Section 5 presents in detail our depth-first branch-and-bound algorithm. Section 6 analyzes the time complexity of our algorithm. Section 7 gives the test results of applying the algorithm to several databases. Section 8 concludes with discussions of future research.

## 2 The Learning Problem

A Bayesian network is a directed acyclic graph $G$ that encodes a joint probability distribution over a set of random variables $U = \{X_1, \ldots, X_n\}$. Each node of the graph $G$ represents a variable in $U$. If there is a directed edge from node $X_i$ to $X_j$, $X_i \longrightarrow X_j$, we say that $X_i$ is a parent of $X_j$. The graph encodes that each variable is independent of its non-descendants given its parents in the graph. In this paper we only consider discrete random variables and we assume that each variable $X_i$ may take on values from a finite set, $\{x_{i1}, \ldots, x_{ir_i}\}$. A network is quantified by a set of conditional probability tables, $P(x_i|pa_i)$, one table for each node-parents family, where we use $PA_i$ to denote the parent set of $X_i$, $x_i$ and $pa_i$ denote an instantiation of values of $X_i$ and $PA_i$, and the probability table enumerates all possible instantiations. A Bayesian network specifies a unique joint probability distribution over the set of random variables $U$ as [Pearl, 1988]:

$$P(X_1, \ldots, X_n) = \prod_{i=1}^{n} P(X_i|PA_i).$$

Assume we are given a training data set, $D = \{u^1, u^2, \ldots, u^N\}$, where each $u^i$ is a particular instantiation over the set of variables $U$. We only consider situations where the data are complete, that is, every variable in $U$ is assigned a value. To learn a Bayesian network from the training database $D$, we use a scoring function to score each network relative to $D$ and then search through possible network structures for the best scored network. The scoring function we use in this paper is based on the MDL principle [Bouckaert, 1994a, Lam and Bacchus, 1994, Suzuki, 1996] and is given by [Friedman and Getoor, 1999]:

$$MDL(G, D) = \sum_i MDL(X_i|PA_i) \quad (1)$$

$$MDL(X_i|PA_i) = H(X_i|PA_i) + \frac{\log N}{2} K(X_i|PA_i) \quad (2)$$

$$H(X_i|PA_i) = -\sum_{x_i, pa_i} N_{x_i, pa_i} \log\left(\frac{N_{x_i, pa_i}}{N_{pa_i}}\right) \quad (3)$$

$$K(X_i|PA_i) = (r_i - 1) \prod_{X_l \in PA_i} r_l \quad (4)$$

where $H(X_i|PA_i)$ is called the empirical entropy term and $\frac{\log N}{2} K(X_i|PA_i)$ the penalty term. $K(X_i|PA_i)$ represents the number of parameters needed to represent $P(X_i|PA_i)$ and $r_i$ is the number of possible states of $X_i$. $N$ is the total number of samples, $N_{x_i, pa_i}$ is the *sufficient statistics* of the variables $X_i$ and $PA_i$ in $D$, that is, the number of samples in $D$ which match the particular instantiation $x_i$ and $pa_i$, and the summation in (3) is over all the possible instantiations of $X_i$ and $PA_i$. We have

$$\sum_{x_i=x_{i1}}^{x_{ir_i}} N_{x_i, pa_i} = N_{pa_i}. \quad (5)$$

When the MDL scoring function is defined as in equations (1)–(4), we wish to find the network structure that *minimizes* the MDL score.

In this paper we assume that a consistent variable ordering is given as $X_1 < X_2 < \ldots < X_n$, where $X_i < X_j$ means that the edge between $X_i$ and $X_j$ can only be directed as $X_i \rightarrow X_j$, not the other way. The scoring function given in equation (1) is *decomposable*, that is, it is decomposed into a sum of local scores over each node-parents family. This decomposability plus node ordering greatly reduces the search complexity. The score $MDL(G, D)$ is minimized if and only if each local score $MDL(X_i|PA_i)$ is individually minimized. Thus each parent set $PA_i$ may be independently selected. The network learning problem reduces to that of for each variable $X_i$ finding a subset of $\{X_1, X_2, \ldots, X_{i-1}\}$ as $PA_i$ that minimizes the score $MDL(X_i|PA_i)$. For each variable $X_i$, we need to search through $2^{i-1}$ sets. For a Bayesian network of $n$ variables, to find the best scored network we need to search through $\sum_{i=1}^{n} 2^{i-1} = 2^n - 1$ sets. The search space is still exponential in the number of variables.

## 3 Previous Work

In this section we examine some previous work that is directly related to the result presented in this paper. Since the possible network structures to search are exponential in the number of variables even if an ordering on the variables is given, [Cooper and Herskovits, 1992] developed a greedy search algorithm called K2 (which used a Bayesian scoring function). To find the parent set for the variable $X_j$, K2 starts with the empty parent set, successively adds to the parent set the variable within $\{X_1, X_2, \ldots, X_{j-1}\}$ that maximally improves a Bayesian score, and stops when no variable can be added such that the score improves. [Bouckaert, 1994a] replaced the Bayesian scoring function in the K2 algorithm with a MDL scoring function and called the resulting algorithm K3. K3 is presented in Figure 1, which will be used for preprocessing in our proposed algorithm.

[Suzuki, 1996] developed an exhaustive search algorithm which uses branch-and-bound (BnB) technique and guarantees to find the parent set with the minimum MDL score. He noticed that in the $MDL$ scoring function

$$MDL(X_j|PA_j) = H(X_j|PA_j) + \frac{\log N}{2} K(X_j|PA_j),$$

when adding a node $X_q$ into $PA_j$, $K(X_j|PA_j)$ increases by $K(X_j|PA_j) * (r_q - 1)$. On the other hand, the empirical entropy term can decrease at most by the current value $H(X_j|PA_j)$ since the value of empirical

**Algorithm K3**

**INPUT**: Node $X_j$;
**OUTPUT**: Parent set $PA_j$, with its MDL score $minMDL$.

$Pred_j = \{X_1, X_2, \ldots, X_{j-1}\}$;
$PA_j = \emptyset$;
$minMDL = MDL(X_j|PA_j)$;
WHILE $(PA_j \neq Pred_j)\{$
   let $X_z$ be the node in $Pred_j \setminus PA_j$ that minimizes $MDL(X_j|PA_j \cup \{X_z\})$;
   $MDL_{new} = MDL(X_j|PA_j \cup \{X_z\})$;
   IF( $MDL_{new} < minMDL$) THEN
     $minMDL = MDL_{new}$;
     $PA_j = PA_j \cup \{X_z\}$;
   ELSE
     RETURN;
$\}$

Figure 1: The K3 algorithm.

entropy is nonnegative. Hence, if we already have

$$H(X_j|PA_j) \leq \frac{\log N}{2} K(X_j|PA_j) * (r_q - 1), \quad (6)$$

adding nodes to the parent set $PA_j$ will always increase the MDL value. Therefore further search along this branch is unnecessary. Based on this observation, Suzuki developed a BnB algorithm termed B&B_D (which is not presented here due to space limit).

Essentially, Suzuki used a heuristic $H(X_j|PA_j) \geq 0$, and found a lower bound for the MDL score:

$$MDL(X_j|PA_j) \geq \frac{\log N}{2} K(X_j|PA_j). \quad (7)$$

In this paper we will improve the B&B_D algorithm in several ways: find better lower bounds for the MDL score, use thus far found minimum MDL score to speed up pruning, and use node ordering etc. But first to clarify the problem we formally define the search space.

## 4 The Search Space

The problem that we are facing is for a variable $X_j$ to find a subset of $U_j = \{X_1, \ldots, X_{j-1}\}$ as its parent set that minimizes the MDL score. To formulate it as a search problem first we define the search space. A search space is composed of a set of states and a set of operators that change the states. A state represents a single configuration of the problem. An operator takes a state and maps it to another state. For our problem a state is naturally defined as a subset of $U_j$ which represents a parent set of $X_j$ and is represented by a set of variables. An operator is defined as adding a single variable to a set of variables.

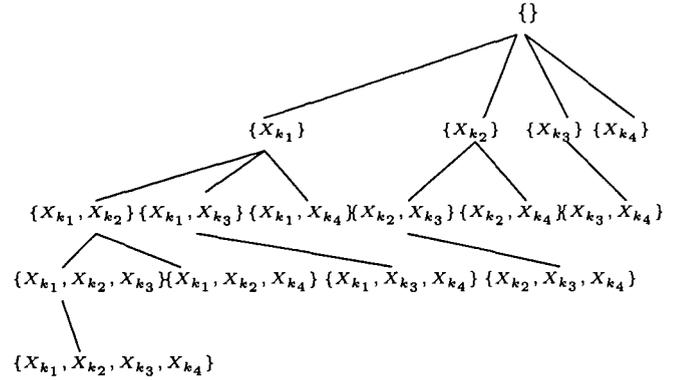

Figure 2: A search tree for finding the parent set of variable $X_5$.

A search space can be represented by a search-space graph. The states are represented by nodes of the graph and the operators by edges between nodes. When a search-space graph is formed as a tree, systematic search algorithms can be applied which are guaranteed to find a solution if one exists. For the current problem we start searching from the empty set, that is, we define the empty set as the root of the search tree. To avoid repeated visit of the same states we give an order to the variables in $U_j$ as

$$X_{k_1} < X_{k_2} < \ldots < X_{k_{j-1}}. \quad (8)$$

We will call this order as *tree order*. The tree order could be arbitrary and has nothing to do with the given variable ordering. We will see later that our algorithm is more efficient under some particular tree order. A state $T$ with $l$ variables is represented by an ordered list $\{X_{i_1}, X_{i_2}, \ldots, X_{i_l}\}$ where $X_{i_1} < X_{i_2} < \ldots < X_{i_l}$ in the tree order. The legal operators for this state $T$ are restricted to adding a single variable that is after $X_{i_l}$ in the tree order. Under the above definitions our search-space graph forms a tree. A search tree for finding the parent set of variable $X_5$ is shown in the Figure 2. The search tree for variable $X_j$ has $2^{j-1}$ nodes and the tree depth is $j - 1$.

For each state $T$ we can compute a MDL score:

$$MDL(X_j|T) = H(X_j|T) + \frac{\log N}{2} K(X_j|T).$$

Systematic search algorithms can be applied to the search tree, which exhaustively search through all states and find the state with the minimum MDL score. We describe our algorithm in the next section.

## 5 Depth-First Branch-and-Bound Algorithm

Inspired by Suzuki's work [Suzuki, 1996] we developed an efficient depth-first branch-and-bound (DFBnB) algorithm. Our algorithm improves Suzuki's B&B_D algorithm in several ways.

1. To determine if a branch of the search tree can be pruned we use thus far found minimum MDL value (denoted by $minMDL$) to compare with the lower bound for the MDL values of that branch.

2. To speed up the pruning, we first run the greedy search algorithm K3 and output the minimum MDL value found by K3 into the DFBnB procedure as the initial $minMDL$ value. Since K3 works very well practically in finding the global minimum MDL value, this leads the DFBnB procedure to prune the search tree nodes using the optimal $minMDL$ value from the starting point most of the time.

3. We find better lower bounds for the MDL values. Assume we are visiting a state $T = \{\ldots, X_{k_l}\}$ in the search tree. Let $W$ denote the set of variables after $X_{k_l}$ in the tree order, $W = \{X_{k_{l+1}}, \ldots, X_{k_{j-1}}\}$. We want to decide if we need to visit the branch below $T$'s child $T \cup \{X_q\}$, where $X_q \in W$. Since the entropy term is nonnegative and $K(X_j|PA_j)$ increases with adding variables to the parent set $PA_j$, Suzuki obtained a lower bound for the MDL values of all states below $T \cup \{X_q\}$ in the search tree as:

$$MDL(X_j|T, X_q, \ldots) \geq \frac{\log N}{2} K(X_j|T) * r_q. \quad (9)$$

Note that all states below $T \cup \{X_q\}$ contain the set $T \cup \{X_q\}$ as a subset. We notice that since the empirical entropy is nonincreasing with respect to adding variables to the parent set [Gallager, 1968], that is,

$$H(X_j|T) \geq H(X_j|T, X_q), \quad (10)$$

all the states below $T$ satisfy

$$H(X_j|T, \ldots) \geq H(X_j|T, W), \quad (11)$$

because they are all subsets of $T \cup W$. Thus a better lower bound for the MDL values of all states below $T \cup \{X_q\}$ is

$$MDL(X_j|T, X_q, \ldots) \geq H(X_j|T, W) + \frac{\log N}{2} K(X_j|T) * r_q. \quad (12)$$

Before computing the entropy term for the state $T \cup \{X_q\}$, if we already have

$$minMDL < H(X_j|T, W) + \frac{\log N}{2} K(X_j|T) * r_q, \quad (13)$$

then all states along the branch below $T \cup \{X_q\}$ can be pruned.

4. We use a node ordering to further speed up the pruning process. So far, the tree order (8) is arbitrary. The entropy terms that are frequently used as lower bounds in equation (12) are (ref. Figure 2): $H(X_j|X_{k_1}, \ldots, X_{k_{j-1}})$, $H(X_j|X_{k_1}, X_{k_3}, \ldots, X_{k_{j-1}})$, $H(X_j|X_{k_2}, \ldots, X_{k_{j-1}})$, $H(X_j|X_{k_3}, \ldots, X_{k_{j-1}})$ etc.. Now the tree order would influence the value distributions of these terms. For example, if the entropy values tend to be lower when the parent sets contain $X_{k_{j-1}}$, all these lower bounds would be small since all these parent sets contain $X_{k_{j-1}}$. On the other hand, if the entropy values tend to be lower when the parent sets contain $X_{k_1}$, most of the lower bounds would be large because most of these parent sets do not contain $X_{k_1}$. The frequencies that the variables appear in these terms are opposite to their tree order, that is, $X_{k_{j-1}}$ appears most frequently and $X_{k_1}$ appears least. Thus naively, if we decide the tree order such that the variables that tend to reduce the entropy are ordered earlier, most of these lower bounds would have larger values than ordered the other way around. In our algorithm, the tree order is determined such that

$$H(X_j|X_{k_1}) \leq H(X_j|X_{k_2}) \leq \ldots \leq H(X_j|X_{k_{j-1}}). \quad (14)$$

Our tests show that our algorithm will visit far fewer states in this order than when the given variable ordering is also used as tree order.

Our proposed algorithm DFBnB_K3 is presented in Figure 3. In the main program we call procedure K3 to find a $minMDL$ value, order the nodes according to equation (14), and call the procedure DFBnB starting from the empty set. The DFBnB procedure is a standard recursive depth-first search algorithm and uses equation (13) as pruning condition.

Some improvement to the DFBnB procedure is possible. The same lower bound terms $H(X_j|T, W)$ may be computed several times. This repeated computation may be avoided by passing their values. Our tests show that this only leads to minor improvement. We used the fixed tree order to visit the search tree. It is possible to dynamically order which branch to visit first according to their MDL values. This way it may be possible to find a lesser $minMDL$ earlier and thus to speed up the pruning. But since we already run K3 procedure which performs very well practically in finding the minimum MDL values, the redundancy of dynamical ordering outweights its possibility of pruning more states. Note that tree order makes the search-space graph form a tree and restricts which variables can be added to a state, but we are still free to choose an order to visit the tree.

**Algorithm DFBnB_K3**

**INPUT**: Node $X_j$;
**OUTPUT**: Parent set $PA_j$, which has the minimum MDL score $minMDL$.

```
main(){
    CALL K3 which outputs minMDL and PA_j;
    order variables according to equation (14);
    p_1 = (log N)/2 (r_j - 1); T = ∅;
    CALL procedure DFBnB(T, X_{k_0}, p_1);
}
```

Procedure DFBnB$(T, X_{k_l}, p_1)${
**INPUT**: state $T$, the last variable $X_{k_l}$ in $T$ in the tree order, penalty term value $p_1$;

$MDL = H(X_j|T) + p_1$;
IF$(MDL < minMDL)$
THEN $minMDL = MDL; PA_j = T$;
Let $W$ be the set of variables after $X_{k_l}$ in tree order;
$lowbound = H(X_j|T, W)$;
FOR(each $X_q \in W$){
  $p_2 = p_1 * r_q$;
  IF $(minMDL > p_2 + lowbound)$
    THEN CALL DFBnB$(T \cup \{X_q\}, X_q, p_2)$;
}
}

Figure 3: Our proposed algorithm DFBnB_K3.

In our algorithm, when we visit a state $T$, we need to compute one extra entropy term $H(X_j|T, W)$ for the use of the lower bound. This computation is costly and at least doubles the time to visit a state. Noticing that

$$H(X_j|T,\ldots) \geq H(X_j|X_1,\ldots,X_{j-1}), \quad (15)$$

we could use a fixed lower bound $H(X_j|X_1,\ldots,X_{j-1})$ in the whole computation, which is still better than using zero as lower bounds. Our tests showed that for small sample size (such as $N = 500$) this could save us time in spite of more states were visited. But for large sample size and number of variables the power of $H(X_j|T,W)$ in pruning states outweights the computation redundancy.

## 6 Complexity Analysis

In this section, we study how many states in the search tree will be visited by our algorithm and thus estimate the time complexity.

Consider a state $T = \{X_{i_1}, X_{i_2}, \ldots, X_{i_{d-1}}\}$ at depth $d-1$ of the search tree. Assume $W$ is the set of variables that are ordered after $X_{i_{d-1}}$ in the tree order. We will expand (i.e., call DFBnB procedure for) a child of $T$, $T \cup \{X_{i_d}\}$, if and only if

$$H(X_j|T,W) + \frac{\log N}{2} K(X_j|T, X_{i_d}) \leq minMDL \quad (16)$$

from the pruning condition (13). Since the number of possible states $r_i \geq 2, i = 1, 2, \ldots, j-1$, we have

$$K(X_j|T, X_{i_d}) \geq (r_j - 1)2^d, \quad (17)$$

thus

$$\frac{\log N}{2}(r_j - 1)2^d \leq minMDL - H(X_j|T,W). \quad (18)$$

$minMDL$ satisfies

$$minMDL \leq MDL(X_j|\emptyset) = H(X_j|\emptyset) + \frac{\log N}{2}(r_j - 1). \quad (19)$$

This leads to

$$\frac{\log N}{2}(r_j - 1)2^d \leq \frac{\log N}{2}(r_j - 1) + H(X_j|\emptyset) - H(X_j|T,W) \quad (20)$$

From information theory we have [Gallager, 1968]

$$H(X_j|\emptyset) \leq N \log r_j \quad (21)$$

and

$$H(X_j|T, W) \geq 0. \quad (22)$$

Thus we obtain

$$\frac{\log N}{2}(r_j - 1)2^d \leq \frac{\log N}{2}(r_j - 1) + N \log r_j, \quad (23)$$

or

$$2^d \leq 1 + \frac{2 \log r_j}{r_j - 1} \frac{N}{\log N}. \quad (24)$$

We conclude that for large $N$, the search tree for a variable $X_j$ is searched at most to the depth of

$$D_j = \lfloor \log \frac{2 \log r_j}{r_j - 1} + \log \frac{N}{\log N} \rfloor, \quad (25)$$

where $\lfloor x \rfloor$ stands for the largest integer not greater than $x$. Our algorithm will not visit parent sets of $X_j$ which contain more than $D_j$ variables. Using the inequality

$$\log r \leq r - 1, \text{ for } r \geq 2, \quad (26)$$

equation (24) becomes

$$2^d \leq 1 + \frac{2N}{\log N}. \quad (27)$$

Thus for any variable our algorithm will not visit parent sets that contain more than

$$D = \lfloor \log \frac{2N}{\log N} \rfloor \quad (28)$$

variables. This is consistent with the theorem in [Bouckaert, 1994b] that the MDL measure will not select network structures that contain parent sets with more than $\log N$ parents. Variable ordering does not influence the result given by equation (28) since it holds under any variable ordering.

**Theorem 1** *The MDL scoring function given in equations (1)–(4) will not select network structures that contain parent sets with more than $\lfloor \log \frac{2N}{\log N} \rfloor$ variables.*

We now estimate how many states the search tree contains to the depth $D_j$. The states at depth $k$ are those subsets of $\{X_1, X_2, \ldots, X_{j-1}\}$ which contain exactly $k$ variables. Thus there are $\binom{j-1}{k}$ states at depth $k$. The total number of states to depth $D_j$ is $m_j = \sum_{k=0}^{D_j} \binom{j-1}{k}$. Assume $D_j \leq (j-1)/2$, we have $m_j < D_j \binom{j-1}{D_j}$. Each state takes $O(N(j-1+r_j)) = O(N(j-1))$ time to compute the entropy terms, where $j-1$ comes from that parent sets can contain at most $j-1$ variables and we assume using hash tables to store the sufficient statistics. Thus the time complexity of finding the minimum scored parent set for variable $X_j$ is

$$T_j = O(N(j-1)D_j \binom{j-1}{D_j}). \qquad (29)$$

For a Bayesian network of $n$ variables, the time complexity to reconstruct the whole structure for large $n$ and $N$ is

$$T(n, N) < n * T_n = O(n^2 N D \binom{n}{D}) \qquad (30)$$

where $D$ is given in equation (28) and we assume $D < \frac{n}{2}$. We may estimate $\binom{n}{D}$ using Stirling's formula $x! \approx \sqrt{2\pi x}(\frac{x}{e})^x$. It can be shown that

$$\binom{n}{D} < \sqrt{\frac{n}{2\pi D(n-D)}}(\frac{en}{D})^D e^{-\frac{1}{2}D(\frac{D}{n})}. \qquad (31)$$

Thus

$$T(n, N) = O(n^2 N \sqrt{D}(\frac{en}{D})^D e^{-\frac{1}{2}D(\frac{D}{n})}) \qquad (32)$$

To see clearly how $T(n, N)$ changes with respect to $N$, using $x^{\log y} = y^{\log x}$ and ignoring the floor function in the expression (28) for $D$, the above formula can also be expressed as

$$T(n, N) = O(n^2 N \sqrt{D}(\frac{2N}{\log N})^{\log \frac{en}{D}} e^{-\frac{1}{2}D(\frac{D}{n})}). \qquad (33)$$

In deriving equation (28) for $D$, we used equation (22), that is, we take zero as the lower bounds for the entropy terms. In practice the values of $H(X_j|T,W)$ used in equation (16) could be large and comparable with $minMDL$ sometimes, especially after we determined the tree order according to equation (14). Thus not all branches are searched to the depth $D$ and some branches are pruned after only a few steps. The fact that many variables have more than two states ($r_i \geq 2$) also leads to that the actual search depth is less than the estimation (28). Our test results in Section 7 show that the DFBnB_K3 algorithm performs far better than the time complexity estimation (33). The actual time grows much slower with respect to $N$. Practically we can get an empirical estimation for $D$ after the K3 procedure has returned a $minMDL$ value. From equation (18) and (15) we have

$$\frac{\log N}{2}(r_j - 1)2^d \leq minMDL - H(X_j|X_1, X_2, \ldots, X_{j-1}),$$

thus a good estimation for $D_j$ is

$$D_j = \lfloor \log \frac{2(minMDL - H(X_j|X_1, X_2, \ldots, X_{j-1}))}{(r_j - 1)\log N} \rfloor. \qquad (34)$$

## 7 Test Results

We applied the DFBnB_K3 algorithm to training data generated from the following networks: ALARM [Beinlich et al., 1989] which contains 37 variables and 46 edges, Boerlage92 [Boerlage, 1992] which has 23 variables and 36 edges, Car_Diagnosis_2 which has 18 variables and 20 edges, Hailfinder2.5 [Abramson et al., 1996] which has 56 variables and 66 edges, **A** [Kozlov and Singh, 1996] which has 54 variables, and **B** [Kozlov and Singh, 1996] which has 18 variables and 39 edges. Boerlage92 and Car_Diagnosis_2 were downloaded in the Netica format from the web site of Norsys Software Corporation, http://www.norsys.com. Hailfinder2.5, **A**, and **B** were obtained with GeNIe modeling environment distribution developed by the Decision Systems Laboratory of the University of Pittsburgh (http://www.sis.pitt.edu/~dsl) and were transformed to Netica format using GeNIe. We used the ALARM database generated by Herskovits [Herskovits, 1991] which contains 20000 cases and which gives a variable ordering that is the same order used by [Cooper and Herskovits, 1992] and [Suzuki, 1996]. A database containing 20000 cases was generated for each other network using a demo version of Netica API developed by Norsys Software Corporation (http://www.norsys.com). In the following experiments by "the sample size is $N$" we mean that the first $N$ cases in the databases were used.

To show the efficiency of the algorithm DFBnB_K3, we compared it with Suzuki's B&B_D using the ALARM database. Table 1 shows our test results. "$k$" column stands for that the computation is up to $X_k$ in the variable ordering. "States" column denotes how many times the entropy terms are computed. For B&B_D algorithm, this number is the same as the number of states visited in the search tree, while for DFBnB_K3 algorithm it is approximately double of the number of states visited. Both algorithms were implemented with C++ language and "Time" column gives the actual run time in a Sun Workstation Ultra5. Table 1 shows that our algorithm indeed improves over Suzuki's B&B_D, and more importantly, the time complexity increases more slowly in the sample size $N$.

Table 2 gives the test results of applying DFBnB_K3 algorithm on various databases with various sample sizes, which lists "States" and "Time" for recovering the full network structures. It shows that the number of states visited increases slowly with sample size $N$, and since the time spent in each state is $O(N)$, the overall time complexity is a low polynomial in $N$. This slow increase rate in $N$ makes it possible to exploit large databases using our algorithm. The table didn't give results for **A** network because it took DFBnB_K3 more than 48 hours to recover the full structure even with $N = 250$. **A** is a randomly generated network originally, has 54 variables and is densely connected.

Since DFBnB_K3 is an exhaustive search algorithm which returns with the *global* minimum MDL score, it paves the way for studying some other problems. One interesting problem is how well the greedy search algorithm K3 does in finding parent sets with the global minimum score. Our test results show that K3 performed very well and found the global minimum most of the time regardless of the sample sizes and databases. For the ALARM database, K3 was trapped in a local minimum twice for $N = 1000$ and once for each other sample size in finding the parent sets for the 36 variables. For the **A** database K3 was trapped 5 times for 54 variables with $N = 250$. For all other databases and sample sizes shown in the Table 2 K3 found the global minimum all the time.

Another interesting problem is how well the MDL scoring function is in recovering the original network structure. Table 3 gives our test results. It lists in the networks with the minimum MDL score how many edges are extra to or missing from the original networks from which the training databases were generated. The results show that the MDL score tends to miss edges when the training data size is small and will add few extra edges even if the data size is small. The number of missing edges decreases as the data size increases, as expected. [Bouckaert, 1994b] has shown that the MDL scoring function will pick the minimal I-map under given variable order when the data size is large enough. The MDL scoring function seems having trouble in recovering structures for **B** network. **B** is a randomly generated network and is densely connected. It has several variables having more than 4 parents but the empty set had the minimum MDL score (thus was picked up as the parent sets) for all of them up to $N = 16000$.

Table 1: Comparison of Suzuki's algorithm with ours (ALARM database).

$N = 250$

| k | B&B_D States | Time (minute) | DFBnB_K3 States | Time (minute) |
|---|---|---|---|---|
| 10 | 649 | 1 | 190 | 1 |
| 15 | 6317 | 1 | 704 | 1 |
| 20 | 26760 | 1 | 2401 | 1 |
| 25 | 57928 | 1 | 4843 | 1 |
| 30 | 104129 | 1 | 6760 | 1 |
| 31 | 154298 | 2 | 9134 | 1 |
| 32 | 165052 | 2 | 11373 | 1 |
| 33 | 197883 | 2 | 13357 | 1 |
| 34 | 215289 | 2 | 13879 | 1 |
| 35 | 236506 | 2 | 14695 | 1 |
| 36 | 264405 | 3 | 15990 | 1 |
| 37 | 324472 | 3 | 17660 | 1 |

$N = 500$

| k | B&B_D States | Time (minute) | DFBnB_K3 States | Time (minute) |
|---|---|---|---|---|
| 10 | 807 | 1 | 220 | 1 |
| 15 | 10350 | 1 | 782 | 1 |
| 20 | 54841 | 1 | 3153 | 1 |
| 25 | 123138 | 2 | 6489 | 1 |
| 30 | 241565 | 4 | 9091 | 1 |
| 31 | 374421 | 7 | 12215 | 1 |
| 32 | 416052 | 7 | 19350 | 2 |
| 33 | 498796 | 9 | 21244 | 2 |
| 34 | 547686 | 9 | 21988 | 2 |
| 35 | 601752 | 10 | 23410 | 2 |
| 36 | 695805 | 12 | 25582 | 2 |
| 37 | 920851 | 16 | 29142 | 2 |

$N = 1000$

| k | B&B_D States | Time (minute) | DFBnB_K3 States | Time (minute) |
|---|---|---|---|---|
| 10 | 908 | 1 | 230 | 1 |
| 15 | 14397 | 1 | 1114 | 1 |
| 20 | 101555 | 4 | 3929 | 1 |
| 25 | 259142 | 9 | 8649 | 1 |
| 30 | 532494 | 18 | 11358 | 2 |
| 31 | 808540 | 28 | 15177 | 2 |
| 32 | 946658 | 33 | 28374 | 4 |
| 33 | 1208414 | 42 | 29912 | 4 |
| 34 | 1372767 | 48 | 30596 | 4 |
| 35 | 1567279 | 54 | 32406 | 4 |
| 36 | 1780786 | 61 | 35674 | 5 |
| 37 | 2423177 | 85 | 39382 | 5 |

Table 2: The efficiency of DFBnB_K3 algorithm. (time is in minutes)

|  | ALARM | | Boerlage92 | | Car_Diagnosis_2 | | B | | Hailfinder2.5 | |
|---|---|---|---|---|---|---|---|---|---|---|
| N | States | Time | States | Time | States | Time | States | Time | States | Time |
| 250 | 17,660 | 1 | 81,695 | 3 | 1,954 | 1 | 2,209 | 1 | 277,492 | 18 |
| 500 | 29,142 | 2 | 111,076 | 7 | 1,724 | 1 | 7,120 | 2 | 621,488 | 72 |
| 1000 | 39,382 | 5 | 169,320 | 19 | 2,731 | 1 | 7,717 | 2 | 1,876,046 | 403 |
| 2000 | 52,047 | 11 | 272,136 | 60 | 2,913 | 1 | 7,725 | 2 | | |
| 4000 | 75,113 | 32 | 320,038 | 148 | 4,026 | 1 | 21,232 | 9 | | |
| 8000 | 112,599 | 93 | 429,900 | 391 | 4,144 | 2 | 21,926 | 15 | | |
| 16000 | 182,627 | 310 | | | 4,376 | 5 | 43,886 | 74 | | |
| nodes | 37 | | 23 | | 18 | | 18 | | 56 | |

Table 3: The performance of MDL scoring function in recovering network structures.

|  | ALARM | | Boerlage92 | | Car_Diagnosis_2 | | B | | Hailfinder2.5 | |
|---|---|---|---|---|---|---|---|---|---|---|
| N | extra | missing | extra | missing | extra | missing | extra | missing | extra | missing |
| 250 | 2 | 15 | 1 | 11 | 0 | 10 | 0 | 38 | 9 | 33 |
| 500 | 0 | 7 | 0 | 12 | 0 | 8 | 0 | 37 | 7 | 26 |
| 1000 | 1 | 5 | 0 | 9 | 0 | 6 | 0 | 35 | 6 | 18 |
| 2000 | 1 | 4 | 0 | 8 | 0 | 5 | 0 | 34 | | |
| 4000 | 0 | 3 | 0 | 8 | 0 | 3 | 0 | 28 | | |
| 8000 | 0 | 1 | 0 | 8 | 0 | 1 | 0 | 25 | | |
| 16000 | 0 | 1 | | | 0 | 1 | 0 | 21 | | |
| edges | 46 | | 36 | | 20 | | 39 | | 66 | |

## 8 Conclusion

We developed an efficient depth-first branch-and-bound search algorithm for learning Bayesian network structures from data when a consistent variable ordering is given. Preliminary test results are promising. The time complexity of our algorithm grows slowly with the sample size. Our algorithm finds the global optimum networks according to the MDL scoring function, thus it can be used to empirically measure the performance of the MDL principle in learning Bayesian network structures. We also showed that the suboptimal algorithm K3 performs very well practically in finding the global optimum.

The branch-and-bound technique is limited to learning based on the MDL principle since it relies on the nature of the MDL scoring function. On the other hand, the technique can be extended to the general cases of not requiring variable ordering, where the greedy search is the common practice. As the DF-BnB_K3 algorithm presented in this paper, we may run a greedy search procedure before a depth-first branch-and-bound procedure. Since the search space is huge, it would be impractical to visit all states even if good pruning is applied. The purpose would be to find a better solution than what greedy search found if more time is spent, since simple greedy search cannot benefit from more time. When we are willing to spend more time, iterative greedy search can be applied though. It would be interesting to compare the performance of branch-and-bound algorithm with the iterative greedy search algorithm. We are currently working on the problem.

### Acknowledgements

The author was supported by a Microsoft Fellowship.